\documentclass[cameraready]{Interspeech}

\usepackage{todonotes}
\usepackage{soul}

\definecolor{TodoColor}{rgb}{1,0.7,0.6}

\usepackage{placeins}
\newcommand{\huggingfacesmall}{\includegraphics[width=9px]{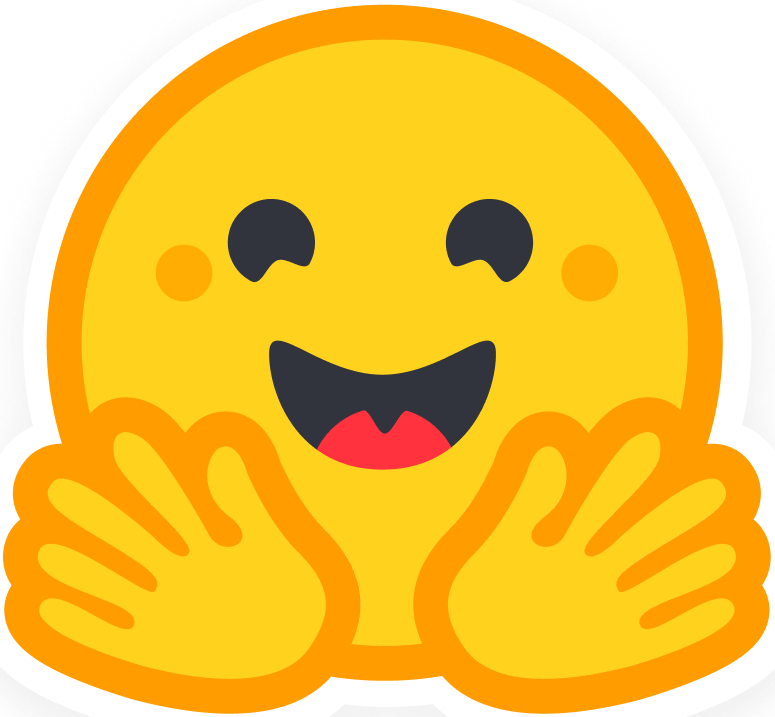}}

\usepackage[capitalise]{cleveref}
\crefname{figure}{Figure}{Figures}
\crefname{table}{Table}{Tables}
\crefname{appendix}{Appendix}{Appendices}

\usepackage{multirow}
\usepackage{subcaption}

\usepackage[table]{xcolor}
\definecolor{taskgreen}{RGB}{180, 220, 180}
\definecolor{taskblue}{RGB}{180, 210, 230}
\definecolor{taskorange}{RGB}{255, 210, 170}
\definecolor{taskpink}{RGB}{255, 185, 200}

\renewcommand{\paragraph}[1]{\medskip\noindent\textbf{#1}\quad}

\usepackage{booktabs}
\usepackage{tabularx}
\usepackage{arydshln}

\makeatletter
\def\adl@drawiv#1#2#3{%
  \hskip.5\tabcolsep
  \xleaders#3{#2.5\@tempdimb #1{1}#2.5\@tempdimb}%
    #2\z@ plus1fil minus1fil\relax
  \hskip.5\tabcolsep
}
\newcommand{\cdashlinelr}[1]{%
  \noalign{\vskip\aboverulesep
           \global\let\@dashdrawstore\adl@draw
           \global\let\adl@draw\adl@drawiv}%
  \cdashline{#1}%
  \noalign{\global\let\adl@draw\@dashdrawstore
           \vskip\belowrulesep}%
}
\makeatother

\title{Do What I Say: A Spoken Prompt Dataset for Instruction-Following}

\author[affiliation={1}, orcid=0009-0001-7238-7705]{Maike}{Züfle}
\author[affiliation={2}, orcid= 0000-0002-4494-8886]{Sara}{Papi}
\author[affiliation={1}, orcid= 0009-0001-4679-0015]{Fabian}{Retkowski}
\author[affiliation={3,4}, orcid= 0009-0006-7557-0157]{Szymon}{Mazurek}
\author[affiliation={3}, orcid= 0000-0002-8473-5236]{Marek}{Kasztelnik}
\author[affiliation={1,5}, orcid= 0009-0001-0848-1000]{Alexander}{Waibel}
\author[affiliation={2}, orcid= 0000-0001-7480-2231]{Luisa}{Bentivogli}
\author[affiliation={1}, orcid= 0000-0002-4231-6543]{Jan}{Niehues}

\address{
    $^1$ Karlsruhe Institute of Technology, Germany \\
    $^2$ Fondazione Bruno Kessler, Italy \\
    $^3$ ACC Cyfronet AGH, Poland \\
    $^4$ AGH University of Krakow, Poland \\
    $^5$ Carnegie Mellon University, U.S.
}

\email{maike.zuefle@kit.edu}

\keywords{benchmark, dataset, instruction following, prompt, spoken prompt, multilingual, crosslingual}

\usepackage{comment}

\begin{document}

\maketitle

\begin{abstract}
Speech Large Language Models (SLLMs) have rapidly expanded, supporting a wide range of tasks. These models are typically evaluated using text prompts, which may not reflect real-world scenarios where users interact with speech. 
To address this gap, we introduce DoWhatISay (DOWIS), a multilingual dataset of human-recorded spoken and written prompts designed to pair with any existing benchmark for realistic evaluation of SLLMs under spoken instruction conditions. 
Spanning 9 tasks and 11 languages, it provides 10 prompt variants per task-language pair, across five styles.
Using DOWIS, we benchmark state-of-the-art SLLMs, analyzing the interplay between prompt modality, style, language, and task type. Results show that text prompts consistently outperform spoken prompts, particularly for low-resource and cross-lingual settings. Only for tasks with speech output, spoken prompts do close the gap, highlighting the need for speech-based prompting in SLLM evaluation.
\end{abstract}

\section{Introduction}
Speech Large Language Models (SLLMs) have seen remarkable progress in recent years, demonstrating strong performance across both speech and text tasks \cite{xu2025qwen3omnitechnicalreport, arora2025landscapespokenlanguagemodels}. A key capability of these models is instruction-following (IF): rather than requiring a special tag or argument to specify a task, they can be guided through natural language prompts \cite{brown_fewshot, su-etal-2023-pandagpt, tang2024salmonn}.
Given this, evaluating how reliably these models follow instructions is critical. However, current speech IF benchmarks 
are mostly conducted with 
text prompts \cite{lu25c_interspeech, yang-etal-2024-air, wang2026mmsumassivemultitaskspoken, gao-etal-2025-benchmarking, huang2025dynamicsuperb, pandey-etal-2025-sift}, an approach that is insufficient for two reasons. First, robust models should handle 
a range of 
prompt styles rather than being evaluated on a single wording \cite{papi2026mcif}. Second, and more importantly, a central motivation for building SLLMs is to enable natural, human-like interaction through speech. Evaluating models on spoken instructions such as ``\textit{Summarise this meeting}'' or ``\textit{Translate what the other person said}'', therefore reflects realistic use cases better than text-only evaluation \cite{chen2024voicebenchbenchmarkingllmbasedvoice, wang2025voiceassistantevalbenchmarkingaiassistants}.

Generating text prompts for speech instruction-following benchmarks is substantially more straightforward: they are often produced using an LLM \cite{yang25g_interspeech, yang-etal-2024-air, wang2026mmsumassivemultitaskspoken, gao-etal-2025-benchmarking, pandey-etal-2025-sift} or written by hand \cite{papi2026mcif}, whereas spoken prompts require human recording, making collection significantly more costly. SpeechInstructBench \cite{wang2025inserterspeechinstructionfollowing} and Uro-Bench \cite{yan-etal-2025-uro} are among the few benchmarks with spoken instructions, but both have notable limitations. First, their instructions are generated using text-to-speech systems, cover only English and Chinese, and are pre-concatenated with task-specific inputs, making them impossible to reuse with other datasets. Second, they focus on general instruction-following and reasoning tasks, whereas researchers also need to evaluate spoken instruction-following for specific tasks such as speech recognition or audio chaptering. Lastly, these benchmarks only support monolingual tasks, and do not target cross-lingual tasks such as speech translation. Consequently, spoken instruction evaluation remains rare and, when it does exist, is mostly limited to the specific case of (monolingual) spoken question answering \cite{lu25c_interspeech, wang2025inserterspeechinstructionfollowing, alam2025spokennativqamultilingualeverydayspoken, chen2024voicebenchbenchmarkingllmbasedvoice}, where spoken questions are also often synthetically generated \cite{wang2025inserterspeechinstructionfollowing, yan-etal-2025-uro}. 

To address this gap, we introduce \underline{\textbf{Do}}\underline{\textbf{W}}hat\underline{\textbf{IS}}ay (DOWIS), the first multilingual prompt dataset with parallel spoken and textual prompts written and recorded by native speakers. Unlike existing benchmarks, DOWIS keeps instructions decoupled from task inputs, so they can be paired with \textit{any} existing benchmark, lowering the barrier to spoken IF evaluation without sacrificing naturalness or linguistic diversity. DOWIS includes nine tasks across speech-to-text (automatic speech recognition, speech QA, audio chaptering, speech translation, speech summarisation), text-to-text (machine translation, text summarization), text-to-speech (text synthesis), and speech-to-speech (speech-to-speech translation), selected to cover the full range of SLLMs input-output modalities from well-studied (e.g., speech recognition and translation) to more complex, less explored tasks (e.g., audio chaptering and summarisation) \cite{retkowski-etal-2025-summarizing}.  DOWIS spans 11 languages (de, en, it, cs, es, fr, hu, nl, pt, ru, sv), totalling 3h17m of audio. 
For each task-language pair, it provides 10 prompt variants spanning five categories: basic, detailed, short, formal, and informal, with two prompts per category.
DOWIS' parallel speech-text design allows analysis of (a) how model performance changes when switching from text to spoken instructions, (b) how performance varies across prompt styles, and (c) the interplay with different languages.

We evaluate two state-of-the-art SLLMs, Phi-4 Multimodal \cite{microsoft2025phi4minitechnicalreportcompact} and Qwen2.5-Omni \cite{xu2025qwen25omnitechnicalreport}, using DOWIS. We find that for tasks with text output, text prompts significantly overestimate the models' performance in contrast to spoken prompts, while for tasks with speech output, such as text-to-speech synthesis or speech-to-speech translation, spoken prompts perform on par or better. Regarding prompt style,  informal text and spoken instructions, such as ``\textit{Hey, can you write out what's being said in this audio?}'', consistently perform worse across tasks. These findings demonstrate that text-based evaluation alone, conducted with limited prompt diversity, paints an overly optimistic picture of model capabilities. This highlights the need for a diverse spoken prompt dataset such as DOWIS,\footnote{DOWIS is available under CC-BY license at \huggingfacesmall{} \href{https://huggingface.co/datasets/maikezu/dowis}{maikezu/dowis} and \url{https://github.com/MaikeZuefle/DOWIS}.} 
which can be easily combined with existing benchmarks to enable more realistic and comprehensive evaluation.

\section{The DOWIS Prompt Dataset}

We introduce DOWIS, the first multilingual speech prompt dataset comprising parallel spoken and textual instructions for nine speech and language processing tasks across 11 languages. DOWIS can be combined with any existing downstream task benchmarks and, therefore, designed to facilitate multifaceted evaluation of instruction-following speech models. When combining DOWIS prompts with a downstream benchmark, the instruction and the audio input will naturally come from different speakers. This reflects realistic usage scenarios: for instance, a user might ask a model ``\textit{Translate what was said in this presentation}'' or ``\textit{Summarise what the other person said in this meeting},'' where the instruction speaker and the audio content speaker are inherently different. 

\paragraph{Prompt Collection and Translation.}
To collect a set of prompts for each task, we ask researchers specialized in the respective tasks to collect two English prompts that they would use in their research as the \textit{basic} prompt and to provide two rephrasings for \textit{formal}, \textit{informal}, \textit{detailed}, and \textit{short} styles, resulting in 10 prompts per task. The styles are defined as follows: \textit{basic} prompts reflect natural, everyday phrasing a researcher would use; \textit{formal} prompts use professional, polished language; \textit{informal} prompts are conversational and casual; \textit{detailed} prompts provide more explicit and precise instructions on how to perform the task; and \textit{short} prompts are as concise as possible while remaining unambiguous.

We collect prompts for nine speech and language tasks: Automatic Speech Recognition (ASR), Text-to-Speech (TTS), Speech Translation (ST), Machine Translation (MT), Speech-to-Speech Translation (S2ST), Speech Summarization (SSUM), Text Summarization (TSUM), Audio Chapter Generation (ACHAP), and Spoken Question Answering (SQA).

We then ask native speakers to translate these prompts from English (en) to 10 languages: German (de), Italian (it), Spanish (es), French (fr), Portuguese (pt), Dutch (nl), Swedish (sv), Czech (cs), Hungarian (hu), and Russian (ru). 
We instruct translators to adapt the prompts so that they would sound \textit{natural} in the target language.
This results in a total of 990 unique text prompts (10 prompts × 9 tasks × 11 languages).

\paragraph{Recording.} 
We ask 19 native or highly proficient language speakers per language to read out the 90 prompts as if giving them to an AI model, using their phone or laptop to simulate a realistic meeting scenario.
We convert all audio recordings to .wav format and trim silence from the start and end of each file using a loudness-based voice activity detection approach. Specifically, we apply a sliding window of 10 ms chunks and mark regions as non-silent if their loudness exceeds $-40$ dBFS. The audio is then cropped to the first and last detected non-silent regions, with a 500 ms padding retained on each side to preserve natural speech onset and offset and avoid abrupt cuts.

\begin{table}[!ht]
\centering
\footnotesize
\begin{tabular}{l@{\hspace{3pt}}p{0.1cm}p{0.1cm}p{0.1cm}p{0.1cm}p{0.1cm}p{0.1cm}p{0.1cm}p{0.1cm}p{0.1cm}p{0.1cm}p{0.1cm}}
\toprule
 \textbf{Recording}  & \textbf{de} & \textbf{en} & \textbf{it} & \textbf{cs} & \textbf{es} & \textbf{fr} & \textbf{hu} & \textbf{nl} & \textbf{pt} & \textbf{ru} & \textbf{sv}  \\
\midrule
\textbf{\# Male  Spks.}      & 2 & 2 & 2 & 1 & 1 & 1 & 1 & 0 & 0 & 1 & 0  \\
\textbf{\# Female Spks.}     & 2 & 2 & 2 & 1 & 1 & 1 & 0 & 1 & 1 & 0 & 1  \\
\midrule
\textbf{Spk. Avg.} \textit{(min)}  & 8 & 8 & 9 & 8 & 9 & 9 & 11 & 10 & 8 & 9 & 8  \\
\textbf{Total}  \textit{(min)}     & 33 & 31 & 35 & 17 & 17 & 18 & 11 & 10 & 8 & 9 & 8  \\
\midrule
\multicolumn{12}{c}{\textbf{9 tasks, 11 langs; Total Audio: 3h17m}}\\
\bottomrule
\end{tabular}
\caption{Annotators per language for DOWIS prompts.}\vspace{-0.5cm}
\label{tab:annotation_overview}
\end{table}

\paragraph{Statistics.}
\cref{tab:annotation_overview} shows the distribution of speakers across languages. In total, we recruited 19 speakers (9 male, 10 female) across 11 languages, including 4 bilingual speakers who recorded in two languages each. On average, each speaker recorded approximately 8m 35s of audio.
\cref{tab:task_durations} shows the average duration per prompt across different tasks. Most task prompts average around 4-5 seconds, with ACHAP being longer at 16 seconds due to its detailed formatting instructions.
When accounting for all speaker recordings across the 11 completed languages, our dataset has a total duration of 3h and 17m.

\begin{table}[t]
    \centering
    \footnotesize
    \begin{tabular}{cccccc}
        \toprule
        \multicolumn{6}{c}{\textbf{Monolingual Tasks }} \\
        \midrule
        ASR  &  SQA & ACHAP & TTS  \\
        4.3s &  5.0s & 15.8s &  4.4s\\
        \midrule
        \multicolumn{6}{c}{\textbf{Monolingual (*) and Crosslingual (\textdagger) Tasks }} \\
        \midrule
        MT \textdagger & ST \textdagger &  TSUM * \textdagger & SSUM * \textdagger  & S2ST \textdagger & \\
        4.1s & 4.5s &  4.6s & 4.6s & 4.8s &\\
        \bottomrule
    \end{tabular}
    \caption{Average duration per prompt across languages.}\vspace{-0.6cm}
    \label{tab:task_durations}
\end{table}
\section{Experiments}

\paragraph{Models.}
We select \texttt{Qwen2.5-Omni-7B}\footnote{\huggingfacesmall{} \href{https://huggingface.co/Qwen/Qwen2.5-Omni-7B}{Qwen/Qwen2.5-Omni-7B}} 
\cite{xu2025qwen25omnitechnicalreport}
and \texttt{Phi-4-multimodal-instruct}\footnote{\huggingfacesmall{} \href{https://huggingface.co/microsoft/Phi-4-multimodal-instruct}{microsoft/Phi-4-multimodal-instruct}} 
\cite{microsoft2025phi4minitechnicalreportcompact}, two state-of-the-art models  (hereafter \textit{Qwen} and \textit{Phi} respectively), to analyze the impact of different prompt types and modalities. Both models are run with default inference parameters and batch size 1 on a single NVIDIA A100-SXM4-40GB GPU.
For each task, we evaluate five prompt styles (\textit{basic}, \textit{formal}, \textit{informal}, \textit{short}, \textit{detailed}) in both text and speech modalities. For all tasks, including cross-lingual tasks, the prompt is formulated in the target language. For tasks requiring speech output (S2ST, TTS), we only evaluate Qwen, as Phi does not support audio generation.

\paragraph{Data.}
We use different datasets across tasks to evaluate model performance under diverse prompt conditions. 
For ASR (\{en, de, it, es, fr, pt, nl, ru, sv, cs, hu\}), MT and ST (en$\rightarrow$\{de, it, es, fr, pt, nl, ru, sv, cs, hu\}), we use FLEURS~\cite{conneau2022fleursfewshotlearningevaluation}. 
For speech-output tasks, we similarly rely on FLEURS for S2ST (\{de, it, es, fr, pt, nl, ru, sv, cs, hu\}$\rightarrow$en) and TTS (en). We restrict outputs to English, as Qwen only supports English speech generation.
For TSUM, SSUM, SQA, and ACHAP, no existing mono- or crosslingual datasets cover all languages. Therefore, for TSUM and SSUM, we use MCIF \cite{papi2026mcif} (en$\rightarrow$\{en, de, it\}), which is also used for SQA (en$\rightarrow$en). Since MCIF provides only written questions, we asked the annotators who recorded the English prompts also to record the MCIF questions, enabling speech-prompt evaluations with spoken questions. 
Finally, we use YTSeg \cite{retkowski2024ytseg} for evaluating the ACHAP (en) task. 

\begin{figure*}[!ht]
    \centering
    \begin{subfigure}[b]{0.49\textwidth}
        \centering
        \includegraphics[width=\textwidth]{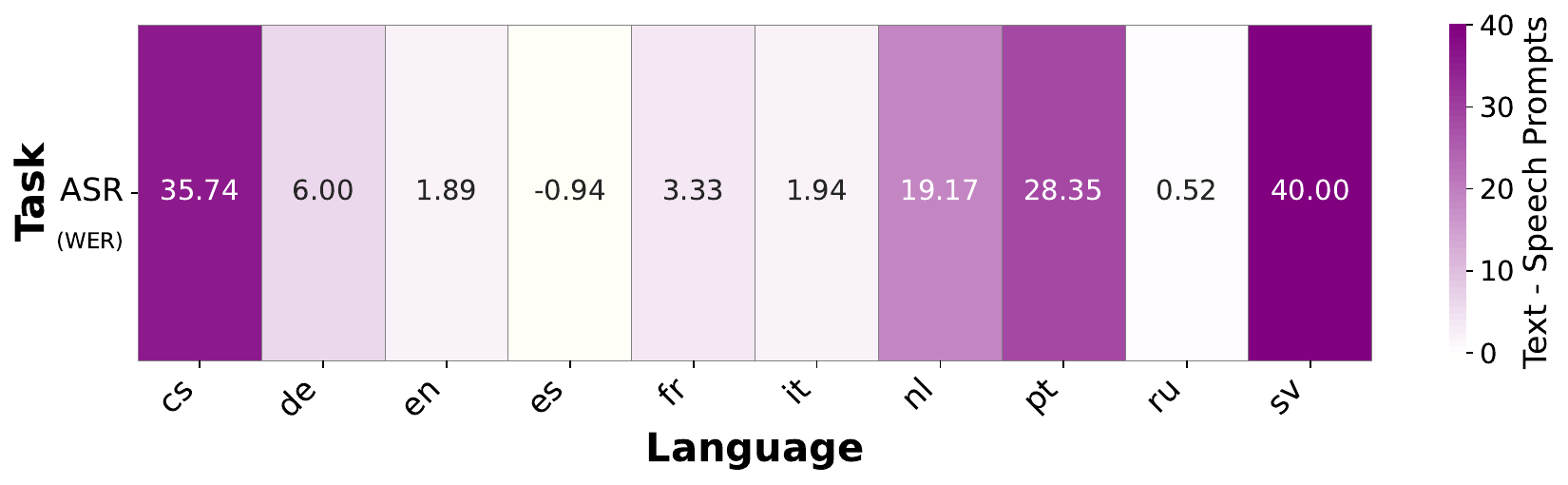}
        \caption{Monolingual Task}
        \label{fig:text_vs_speech_qwen-mono}
    \end{subfigure}
    \hfill
    \begin{subfigure}[b]{0.49\textwidth}
        \centering
        \includegraphics[width=\textwidth]{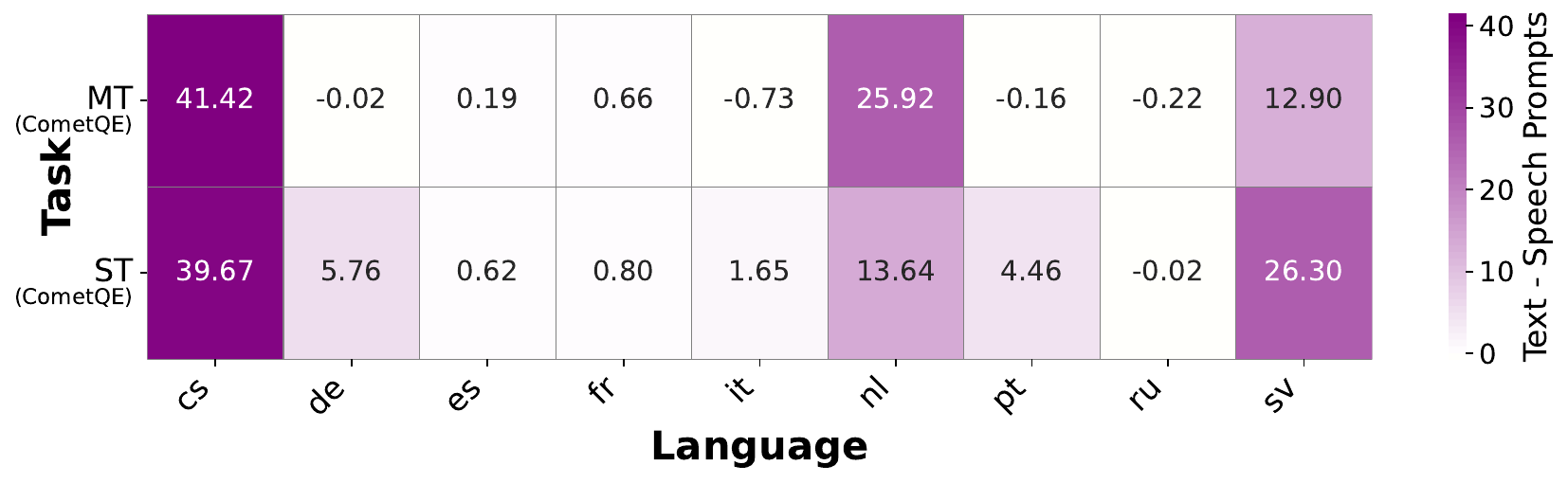}
        \caption{Crosslingual Tasks}
        \label{fig:qwen-multi}
    \end{subfigure}
    \caption{Performance comparison for Qwen: Text Prompt vs Speech Prompts with respect to different target languages. Positive values (purple) indicate text prompt performs better, negative values indicate speech prompts perform better.}\vspace{-0.2cm}
    \label{fig:ext_vs_speech_qwen-heatmaps}
\end{figure*}

\paragraph{Evaluation.}
We evaluate all tasks using standard metrics on the given datasets. Specifically, we evaluate ASR using Word Error Rate (WER) computed with \texttt{jiwer} \cite{wer-jiwer}. For MT and ST, we use CometKiwi \cite{rei-etal-2022-cometkiwi}, a quality estimation metric that does not require reference translations that has shown good correlation with humans for speech and text translation \cite{freitag-etal-2024-llms,papi2025hearingtranslateeffectivenessspeech}. For SQA, SSUM, and TSUM, we follow the evaluation protocol from MCIF \cite{papi2026mcif}, using normalized BERTScore \cite{bert_score} with the \texttt{deberta-xlarge-mnli}\footnote{\huggingfacesmall{} \href{https://huggingface.co/microsoft/deberta-xlarge-mnli}{microsoft/deberta-xlarge-mnli}} \cite{he2021deberta} model to measure semantic similarity between generated and reference answers.
For speech-output tasks (TTS and S2ST), we first transcribe the generated audio using \texttt{whisper-large-v3}\footnote{\huggingfacesmall{} \href{https://huggingface.co/openai/whisper-large-v3}{openai/whisper-large-v3}} \cite{radford2022whisper}. We then report WER for TTS and CometKiwi for S2ST to evaluate content accuracy. To assess speech quality, we additionally report UTMOS \cite{saeki22c_interspeech} for both tasks.
Finally, for ACHAP, we use \texttt{chunkseg} with the evaluation methodology from \cite{retkowski2026transcriptsrenewedperspectiveaudio}, reporting Collar-F1 ($\pm$3s) and BERTScore$_{\text{GC}}$ (Global Concatenation; BERTScore on concatenated predicted vs. reference titles).


\begin{table}[t]
    \centering
    \resizebox{\linewidth}{!}{
    \footnotesize
    \begin{tabular}{p{0.6cm}p{1.3cm}p{0.6cm}p{0.7cm}p{0.7cm}p{0.7cm}p{0.7cm}}
     \toprule
     \multirow{2}{*}{\textbf{Task}} & \multirow{2}{*}{\textbf{Metric}} & \multirow{2}{*}{\textbf{Model}} & \textbf{Text} & \textbf{Speech} & \multicolumn{2}{c}{\textbf{Speech Prompt}} \\
     &&& \textbf{Prompt} & \multicolumn{1}{c}{\textbf{Prompt}} & \multicolumn{1}{c}{\textbf{Male}} & \multicolumn{1}{c}{\textbf{Fem.}} \\
     \midrule
\multirow{2}{*}{\textbf{ASR}} & \multirow{2}{*}{WER $\downarrow$} & Phi* & \cellcolor{taskgreen!79} 16.69 & \cellcolor{taskgreen!20} 332.41 & \cellcolor{taskorange!20} 402.77 & \cellcolor{taskorange!40} 271.15 \\
      &  & Qwen & \cellcolor{taskgreen!80} 12.60 & \cellcolor{taskgreen!79} 17.08 & \cellcolor{taskorange!80} 13.77 & \cellcolor{taskorange!80} 14.91 \\
     \cmidrule(lr){1-7}
     \multirow{2}{*}{\textbf{SQA}} & \multirow{2}{*}{BERTS. $\uparrow$} & Phi & \cellcolor{taskblue!80} 36.49 & \cellcolor{taskblue!20} 11.16 & \cellcolor{taskpink!22} 11.37 & \cellcolor{taskpink!20} 10.96 \\
      &  & Qwen & \cellcolor{taskblue!60} 27.85 & \cellcolor{taskblue!58} 27.12 & \cellcolor{taskpink!80} 27.13 & \cellcolor{taskpink!80} 27.11 \\
     \cmidrule(lr){1-7}
     \multirow{4}{*}{\textbf{ACHAP}} & \multirow{2}{*}{CollarF1 $\uparrow$} & Phi & \cellcolor{taskgreen!28} 7.49 & \cellcolor{taskgreen!20} 6.27 & \cellcolor{taskorange!25} 6.56 & \cellcolor{taskorange!20} 5.99 \\
      &  & Qwen & \cellcolor{taskgreen!80} 15.20 & \cellcolor{taskgreen!66} 13.08 & \cellcolor{taskorange!80} 13.21 & \cellcolor{taskorange!78} 12.96 \\
      \cdashlinelr{2-7}
      & \multirow{2}{*}{GC-BS $\uparrow$} & Phi & \cellcolor{taskgreen!80} 84.30 & \cellcolor{taskgreen!20} 62.43 & \cellcolor{taskorange!40} 64.08 & \cellcolor{taskorange!20} 60.78 \\
      &  & Qwen & \cellcolor{taskgreen!53} 74.28 & \cellcolor{taskgreen!42} 70.50 & \cellcolor{taskorange!76} 70.13 & \cellcolor{taskorange!80} 70.86 \\
     \cmidrule(lr){1-7}

     \multirow{2}{*}{\textbf{TTS}} & UTMOS $\uparrow$ & Qwen & \cellcolor{taskblue!20} 4.33 & \cellcolor{taskblue!80} 4.34 & \cellcolor{taskpink!80} 4.34 & \cellcolor{taskpink!80} 4.34 \\
     \cdashlinelr{2-7}
      & WER$_{\text{ASR}}$ $\downarrow$ & Qwen & \cellcolor{taskblue!20} 30.14 & \cellcolor{taskblue!80} 28.53 & \cellcolor{taskpink!20} 30.64 & \cellcolor{taskpink!80} 26.42 \\
     \midrule
     \multirow{2}{*}{\textbf{MT}} & \multirow{2}{*}{Comet $\uparrow$} & Phi & \cellcolor{taskgreen!73} 77.23 & \cellcolor{taskgreen!20} 46.99 & \cellcolor{taskorange!20} 51.16 & \cellcolor{taskorange!20} 50.96 \\
      &  & Qwen & \cellcolor{taskgreen!80} 81.41 & \cellcolor{taskgreen!62} 70.97 & \cellcolor{taskorange!76} 73.81 & \cellcolor{taskorange!80} 75.41 \\
     \cmidrule(lr){1-7}
     \multirow{2}{*}{\textbf{ST}} & \multirow{2}{*}{Comet $\uparrow$} & Phi & \cellcolor{taskblue!68} 75.82 & \cellcolor{taskblue!20} 57.79 & \cellcolor{taskpink!20} 64.46 & \cellcolor{taskpink!23} 64.82 \\
      &  & Qwen & \cellcolor{taskblue!80} 80.21 & \cellcolor{taskblue!49} 68.57 & \cellcolor{taskpink!67} 71.11 & \cellcolor{taskpink!80} 72.96 \\
     \cmidrule(lr){1-7}
     \multirow{2}{*}{\textbf{TSUM}} & \multirow{2}{*}{BERTS. $\uparrow$} & Phi & \cellcolor{taskgreen!80} 45.81 & \cellcolor{taskgreen!20} 40.91 & \cellcolor{taskorange!28} 41.12 & \cellcolor{taskorange!20} 40.71 \\
      &  & Qwen & \cellcolor{taskgreen!63} 44.44 & \cellcolor{taskgreen!51} 43.41 & \cellcolor{taskorange!80} 43.88 & \cellcolor{taskorange!62} 42.93 \\
     \cmidrule(lr){1-7}
     \multirow{2}{*}{\textbf{SSUM}} & \multirow{2}{*}{BERTS. $\uparrow$} & Phi & \cellcolor{taskblue!65} 43.21 & \cellcolor{taskblue!20} 35.68 & \cellcolor{taskpink!20} 35.59 & \cellcolor{taskpink!21} 35.78 \\
      &  & Qwen & \cellcolor{taskblue!80} 45.66 & \cellcolor{taskblue!65} 43.22 & \cellcolor{taskpink!80} 43.51 & \cellcolor{taskpink!76} 42.92 \\
     \cmidrule(lr){1-7}
    \multirow{2}{*}{\textbf{S2ST}} & UTMOS $\uparrow$ & Qwen & \cellcolor{taskgreen!80} 4.35 & \cellcolor{taskgreen!80} 4.35 & \cellcolor{taskorange!80} 4.36 & \cellcolor{taskorange!80} 4.36 \\
    \cdashlinelr{2-7}
      & COM.$_{\text{ASR}}$$\uparrow$ & Qwen & \cellcolor{taskgreen!80} 72.10 & \cellcolor{taskgreen!20} 72.08 & \cellcolor{taskorange!20} 75.99 & \cellcolor{taskorange!80} 76.18 \\
 
     \bottomrule
    \end{tabular}}    \caption{The impact of speech vs. text prompts. The results are averaged over different prompt types and languages. *Phi only supports the languages 'en', 'de', 'fr', 'it', 'es', 'pt' for speech input, so we limit the evaluation to these for ASR. For the last two columns, only languages are considered, where DOWIS provides male and female speakers (details in \cref{tab:annotation_overview}).}\vspace{-0.6cm}
    \label{tab:modality}
\end{table}

\section{Analysis}
We analyse models' performance on the DOWIS prompts along two dimensions: (1) how well state-of-the-art SLLMs perform on speech vs. text prompts (\cref{subsec:res-txt-vs-speech}), and (2) how much prompt types influence their performance (\cref{subsec:res-prompt-types}).

\subsection{Impact of Text vs. Speech Prompts}
\label{subsec:res-txt-vs-speech}

\paragraph{General Trends.} 
\cref{tab:modality} gives an overview of model performance on text and speech prompts for all nine tasks, averaged over languages and prompt types, with blue and green colours comparing text vs. speech prompts, and orange and pink colours comparing prompts recorded by male vs. female speakers. For tasks with text output, text instructions consistently lead to better results. Only for TSUM and SQA with Qwen, the differences are small. For Phi, speech prompts sometimes lead to severe failure cases: for instance, the WER for ASR exceeds 100, indicating the model cannot handle spoken instructions for this task, despite performing comparably to Qwen when using text prompts. Similar trends are observed on MT and ST. For tasks with speech output (S2ST, TTS) speech and text prompts perform similarly, with speech prompts sometimes performing slightly better.
\begin{figure*}[t]
    \centering
    \begin{subfigure}[b]{0.49\textwidth}
        \centering
        \includegraphics[width=\textwidth]{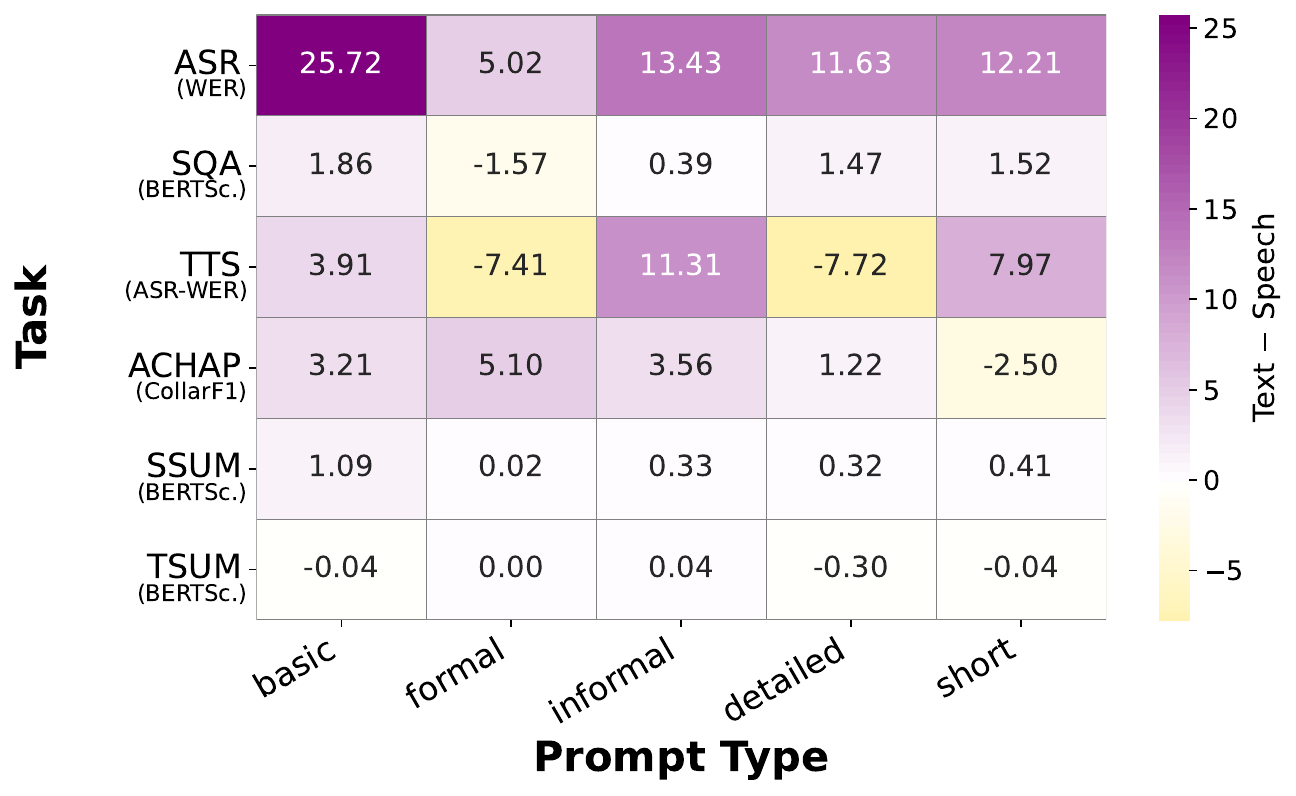}
        \caption{Monolingual Tasks}
        \label{fig:text_vs_speech_qwen-mono_prompt_types}
    \end{subfigure}
    \hfill
    \begin{subfigure}[b]{0.49\textwidth}
        \centering
        \includegraphics[width=\textwidth]{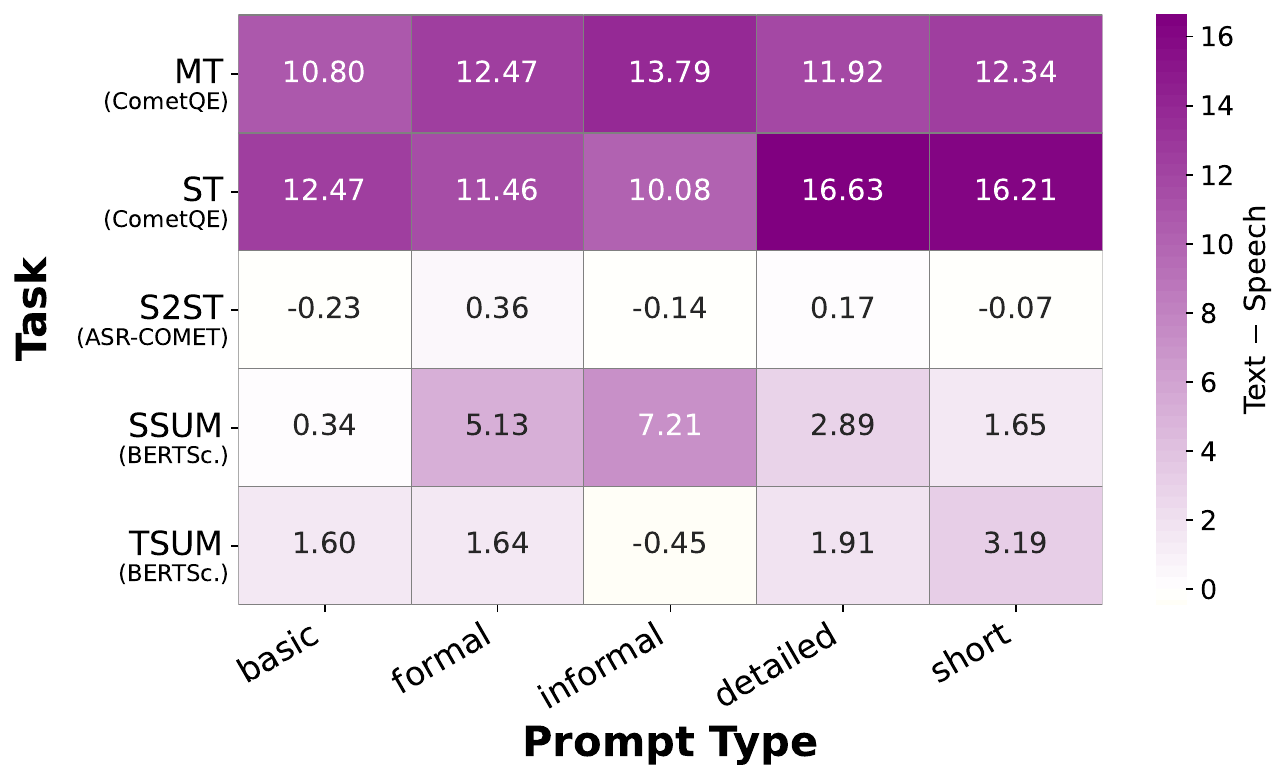}
        \caption{Crosslingual Tasks}
        \label{fig:qwen-multi_prompt_types}
    \end{subfigure}
    \caption{Performance comparison for Qwen2.5-Omni: Text Prompt vs Speech Prompts with respect to different prompt types. Positive values (purple) indicate text prompt performs better, negative values (yellow) indicate speech prompts perform better.}
    \label{fig:prompt_types_heatmap}
\end{figure*}


\begin{table*}
    \centering
    \footnotesize
    \begin{tabular}{p{0.7cm}l ccccc | p{0.6cm}l ccccc}
     \toprule
     \multirow{2}{*}{\textbf{Task}} & \multirow{2}{*}{\textbf{Model}} & \multicolumn{5}{c|}{\textbf{Prompt Type}} &
     \multirow{2}{*}{\textbf{Task}} & \multirow{2}{*}{\textbf{Model}} & \multicolumn{5}{c}{\textbf{Prompt Type}} \\
     \cmidrule(lr){3-7} \cmidrule(lr){10-14}
     && \textbf{Basic} & \textbf{Formal} & \textbf{Inform.} & \textbf{Detail.} & \textbf{Short} &
     && \textbf{Basic} & \textbf{Formal} & \textbf{Inform.} & \textbf{Detail.} & \textbf{Short} \\
     \midrule
     \multirow{2}{*}{\textbf{ASR}$\downarrow$}
&Phi* & \cellcolor{taskgreen!0} 187.49 & \cellcolor{taskgreen!0} 199.13 & \cellcolor{taskgreen!0} 274.56 & \cellcolor{taskgreen!0} 173.05 & \cellcolor{taskgreen!0} 251.39
       & \multirow{2}{*}{\textbf{MT}$\uparrow$}
       & Phi   & \cellcolor{taskgreen!0} 60.35  & \cellcolor{taskgreen!0} 55.14  & \cellcolor{taskgreen!0} 59.62  & \cellcolor{taskgreen!0} 62.44  & \cellcolor{taskgreen!0} 56.86 \\
     & Qwen & \cellcolor{taskgreen!30} 19.79 & \cellcolor{taskgreen!80} 10.51 & \cellcolor{taskgreen!49} 16.32 & \cellcolor{taskgreen!64} 13.39 & \cellcolor{taskgreen!20} 21.58 
     & & Qwen  & \cellcolor{taskgreen!80} 76.55 & \cellcolor{taskgreen!56} 74.75 & \cellcolor{taskgreen!46} 73.98 & \cellcolor{taskgreen!67} 75.56 & \cellcolor{taskgreen!20} 72.08 \\
     \cmidrule(lr){1-7} \cmidrule(lr){8-14}
     \multirow{2}{*}{\textbf{SQA}$\uparrow$}
       & Phi   & \cellcolor{taskblue!0} 20.48  & \cellcolor{taskblue!0} 21.17  & \cellcolor{taskblue!0} 17.58  & \cellcolor{taskblue!0} 21.04  & \cellcolor{taskblue!0} 17.77
       & \multirow{2}{*}{\textbf{ST}$\uparrow$}
       & Phi   & \cellcolor{taskblue!0} 65.65  & \cellcolor{taskblue!0} 65.33  & \cellcolor{taskblue!0} 62.71  & \cellcolor{taskblue!0} 61.19  & \cellcolor{taskblue!0} 64.72 \\
     & Qwen  & \cellcolor{taskblue!80} 28.55 & \cellcolor{taskblue!62} 27.92 & \cellcolor{taskblue!42} 27.26 & \cellcolor{taskblue!23} 26.60 & \cellcolor{taskblue!20} 26.49
     & & Qwen  & \cellcolor{taskblue!71} 73.90 & \cellcolor{taskblue!80} 74.67 & \cellcolor{taskblue!66} 73.49 & \cellcolor{taskblue!20} 69.51 & \cellcolor{taskblue!49} 72.00 \\
     \cmidrule(lr){1-7} \cmidrule(lr){8-14}
     \multirow{4}{*}{\textbf{ACHAP}}
       & Phi$^a$   & \cellcolor{taskgreen!0} 7.66   & \cellcolor{taskgreen!0} 8.48   & \cellcolor{taskgreen!0} 6.95   & \cellcolor{taskgreen!0} 8.11   & \cellcolor{taskgreen!0} 2.18
       & \multirow{2}{*}{\textbf{TSUM}$\uparrow$}
       & Phi   & \cellcolor{taskgreen!32} 40.11 & \cellcolor{taskgreen!69} 44.67 & \cellcolor{taskgreen!62} 43.73 & \cellcolor{taskgreen!73} 45.15 & \cellcolor{taskgreen!24} 39.06 \\
     & Qwen$^a$  & \cellcolor{taskgreen!79} 15.49 & \cellcolor{taskgreen!80} 15.57 & \cellcolor{taskgreen!45} 12.55 & \cellcolor{taskgreen!73} 15.00 & \cellcolor{taskgreen!20} 10.33
     & & Qwen  & \cellcolor{taskgreen!80} 45.98 & \cellcolor{taskgreen!64} 44.07 & \cellcolor{taskgreen!20} 38.62 & \cellcolor{taskgreen!70} 44.77 & \cellcolor{taskgreen!75} 45.31 \\
     \cmidrule(lr){8-14}
     & Phi$^b$   & \cellcolor{taskgreen!75} 79.74 & \cellcolor{taskgreen!78} 81.19 & \cellcolor{taskgreen!54} 69.46 & \cellcolor{taskgreen!84} 84.22 & \cellcolor{taskgreen!0} 34.00
       & \multirow{2}{*}{\textbf{SSUM}$\uparrow$}
       & Phi   & \cellcolor{taskblue!0} 38.54  & \cellcolor{taskblue!0} 38.51  & \cellcolor{taskblue!0} 38.40  & \cellcolor{taskblue!3} 39.48  & \cellcolor{taskblue!0} 36.04 \\
     & Qwen$^b$  & \cellcolor{taskgreen!80} 82.15 & \cellcolor{taskgreen!75} 79.59 & \cellcolor{taskgreen!41} 63.52 & \cellcolor{taskgreen!76} 80.19 & \cellcolor{taskgreen!20} 53.36
     & & Qwen  & \cellcolor{taskblue!80} 45.74 & \cellcolor{taskblue!64} 44.46 & \cellcolor{taskblue!20} 40.87 & \cellcolor{taskblue!64} 44.44 & \cellcolor{taskblue!67} 44.66 \\
     \cmidrule(lr){1-7} \cmidrule(lr){8-14}
     \multirow{2}{*}{\textbf{TTS}}
       & Qwen$^c$ & \cellcolor{taskblue!31} 4.33 & \cellcolor{taskblue!80} 4.35 & \cellcolor{taskblue!26} 4.33 & \cellcolor{taskblue!20} 4.33 & \cellcolor{taskblue!47} 4.34
       & \multirow{2}{*}{\textbf{S2ST}}
       & Qwen$^e$ & \cellcolor{taskgreen!69} 4.35 & \cellcolor{taskgreen!80} 4.35 & \cellcolor{taskgreen!20} 4.34 & \cellcolor{taskgreen!44} 4.35 & \cellcolor{taskgreen!54} 4.35 \\
     & Qwen$^d$ &  \cellcolor{taskblue!80} 8.06 & \cellcolor{taskblue!35} 38.80 & \cellcolor{taskblue!20} 49.31 & \cellcolor{taskblue!42} 34.28 & \cellcolor{taskblue!70} 14.88 
     & & Qwen$^f$ & \cellcolor{taskgreen!28} 71.90 & \cellcolor{taskgreen!42} 72.04 & \cellcolor{taskgreen!20} 71.82 & \cellcolor{taskgreen!71} 72.29 & \cellcolor{taskgreen!80} 72.39 \\
     \bottomrule
    \end{tabular}
    \caption{The impact of prompt typs.  Results are averaged over prompt modalities and languages. Metrics are the same as in \cref{tab:modality}. For tasks with multiple metrics, the metrics are: $^a$CollarF1$\uparrow$, $^b$BERTS.$_{\text{GC}}$$\uparrow$; $^c$UTMOS$\uparrow$, $^d$WER$_{\text{ASR}}$$\downarrow$; $^e$UTMOS$\uparrow$, $^f$COM.$_{\text{ASR}}$$\uparrow$. *Phi only supports the languages 'en', 'de', 'fr', 'it', 'es', 'pt' for speech input, so we limit the evaluation to these for ASR.}\vspace{-0.5cm}
    \label{tab:prompt_type}
\end{table*}
\paragraph{Female vs. Male Audio Prompts.}  For some tasks, models show a small preference for prompts from female or male speakers, though this is not consistent across tasks (for languages where both genders recorded all prompts, ensuring a fair comparison). For TSUM and SSUM, Qwen achieves better results with male prompts, while for TTS, MT, ST, and S2ST, female prompts are preferred. To verify that the observed differences are not driven by acoustic variation in the recordings, we transcribe all audio prompts using  \texttt{whisper-large-v3}\footnote{\huggingfacesmall{} \href{https://huggingface.co/openai/whisper-large-v3}{openai/whisper-large-v3}} \cite{radford2022whisper} and compute WER against the reference prompt texts. With generally high intelligibility across all prompts, we find no clear pattern between prompt WER and model performance, for example, TSUM prompts have similarly low transcription WER for both genders (12\% for both), yet a performance gap persists (BERTScore 43.88 vs.\ 42.93). This suggests that prompt intelligibility is likely not the primary driver of the observed gender differences, which may instead reflect speaker-related biases in the models, as attested by related literature \cite{attanasio-etal-2024-twists,papi2025hearingtranslateeffectivenessspeech}. This highlights the importance of evaluating both genders to detect and address such biases during model development.

\paragraph{Speech Prompts' Interplay with Languages.} To analyse the impact of speech prompts across languages, we focus on Qwen, as it performed better overall, has wider language coverage than Phi, and does not exhibit the severe speech prompt failures discussed above. \cref{fig:ext_vs_speech_qwen-heatmaps} shows the absolute performance difference between text and speech prompts per language for the three tasks covering all languages in DOWIS: ASR, MT, and ST.  Notably, for Czech (cs), Dutch (nl), Portuguese (pt), and Swedish (sv), we observe a strong preference for text instructions in ASR; for MT and ST, a similar pattern holds for cs, nl, and sv.
Looking more closely at the individual results, for ASR in cs and sv, Qwen already achieves a WER above 100 with text prompts, suggesting near-random performance regardless of prompt modality. However, for nl and pt, text WER is between 31--37, and for MT and ST across these languages, text prompts yield COMET scores between 76--82. This suggests that for these languages and tasks, models are capable when using text prompts but struggle to generalise to spoken instructions.
Since participants recorded in realistic conditions using personal devices, we verify prompt intelligibility using the \texttt{whisper-large-v3} transcriptions described above. The overall WER of 12.72\% across all prompts (and per-language WERs of 16\%, 26\%, 18\%, and 13\% for cs, nl, pt, and sv), confirm that the observed performance drop is attributable to the model's difficulty following spoken instructions rather than acoustic variation in the recordings.

\subsection{Impact of Different Prompt Types}
\label{subsec:res-prompt-types}

\paragraph{General Trends.}
\cref{tab:prompt_type} gives an overview of model performance across different prompt types, averaged over prompt modalities and languages. While results vary by task, a clear trend emerges: informal and short prompts are consistently the most challenging across all tasks. Formal and detailed prompts generally perform well, suggesting that models respond better to more structured and explicit instructions.

\paragraph{Speech Prompts' Interplay with Prompt Types.} To examine whether there is an interplay between prompt type and prompt modality, we present \cref{fig:prompt_types_heatmap}, which shows the difference between text and speech prompts for each prompt type and task. For ASR, MT, and ST, text prompts consistently outperform speech prompts across all prompt types. Interestingly, for TTS, we observe a more nuanced pattern: formal and detailed prompts perform better with speech, while basic, informal, and short instructions perform better with text.

\section{Conclusion}
We introduced DOWIS, the first human-recorded parallel spoken-textual prompt dataset for evaluating spoken instruction-following in SLLMs. Covering nine tasks, 11 languages, and five prompt styles, DOWIS can be easily combined with any task-specific benchmark to enable more realistic and comprehensive instruction-following SLLMs evaluation.
Our analysis of Phi-4 Multimodal and Qwen2.5-Omni reveals that current models struggle with spoken instructions for tasks with text output, while performing comparably on tasks with speech output. We further find that informal prompts consistently underperform across tasks, likely due to their more colloquial nature, and that model performance varies across different prompt types and speaker gender. Together, these results demonstrate that text-based evaluation alone paints an overly optimistic picture of model capabilities, and that prompt style, modality, and language all play an important role in instruction-following performance. We hope DOWIS will serve as a useful resource for the community, enabling a multifaceted SLLMs evaluation.

\section{Generative AI Use Disclosure}

Claude was employed exclusively to correct grammar in content authored by humans and in writing code to design paper's plots.

\section{Acknowledgements}
This work has received funding from the European Union’s Horizon research and innovation programme under grant agreement No 101135798, project Meetween (My Personal AI Mediator for Virtual MEETtings BetWEEN People). This research is also supported by the project “How is AI Changing Science? Research in the Era of Learning Algorithms” (HiAICS), funded by the Volkswagen Foundation. We gratefully acknowledge Polish high-performance computing infrastructure PLGrid (HPC Center: ACK Cyfronet AGH) for providing computer facilities and support within computational grant no. PLG/2025/019083.

\bibliographystyle{IEEEtran}
\bibliography{mybib}

\end{document}